\pdfoutput=1

\documentclass[11pt]{article}

\usepackage{ACL2023}

\usepackage{times}
\usepackage{latexsym}
\usepackage[T1]{fontenc}
\usepackage[utf8]{inputenc}
\usepackage{microtype}
\usepackage{inconsolata}
\usepackage{graphicx}
\usepackage{amsmath}
\usepackage{booktabs}
\usepackage{tabularray}
\usepackage{cellspace}
\newcommand\cincludegraphics[2][]{\raisebox{-0.3\height}{\includegraphics[#1]{#2}}}
\usepackage{setspace}
\usepackage{twemojis}

\title{Less is More: Pre-Training Cross-Lingual Small-Scale Language Models with Cognitively-Plausible Curriculum Learning Strategies}

\author{
    {\bf Suchir Salhan} \texttwemoji{lemon}\texttwemoji{mandarin} ~~~~
    {\bf Richard Diehl Martinez} \texttwemoji{lemon} ~~~~
    {\bf Z\'ebulon Goriely} \texttwemoji{lemon} ~~~~
    {\bf Paula Buttery} \texttwemoji{lemon}\texttwemoji{mandarin} ~~~~ \\
    \texttwemoji{lemon} Department of Computer Science \& Technology, University of Cambridge, U.K. \\
    \texttwemoji{mandarin} ALTA Institute, University of Cambridge, U.K. \\
    \texttt{ \{sas245,rd654,zg258,pjb48\}@cam.ac.uk}}

\begin{document}
\maketitle
\begin{abstract}

Curriculum Learning has been a popular strategy to improve the cognitive plausibility of Small-Scale Language Models (SSLMs) in the BabyLM Challenge. However, it has not led to considerable improvements over non-curriculum models. We assess whether theoretical linguistic acquisition theories can be used to specify more fine-grained curriculum learning strategies, creating age-ordered corpora of Child-Directed Speech for four typologically distant language families to implement SSLMs and acquisition-inspired curricula cross-lingually. Comparing the success of three objective curricula (\textsc{Growing, Inwards} and \textsc{MMM}) that precisely replicate the predictions of acquisition theories on a standard SSLM architecture, we find fine-grained acquisition-inspired curricula can outperform non-curriculum baselines and performance benefits of curricula strategies in SSLMs can be derived by specifying fine-grained language-specific curricula that precisely replicate language acquisition theories.

\begin{tblr}{colspec = {Q[c,m]X[l,m]}, stretch = 0}
    \cincludegraphics[width=1.2em, keepaspectratio]{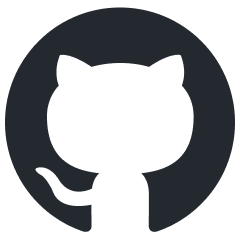} & {\footnotesize{\url{https://github.com/suchirsalhan/MAO-CLIMB} \tiny{(CC BY 4.0)}}}
\end{tblr}
\begin{tblr}{colspec = {Q[c,m]X[l,m]}, stretch = 0}
    \cincludegraphics[width=1.2em, keepaspectratio]{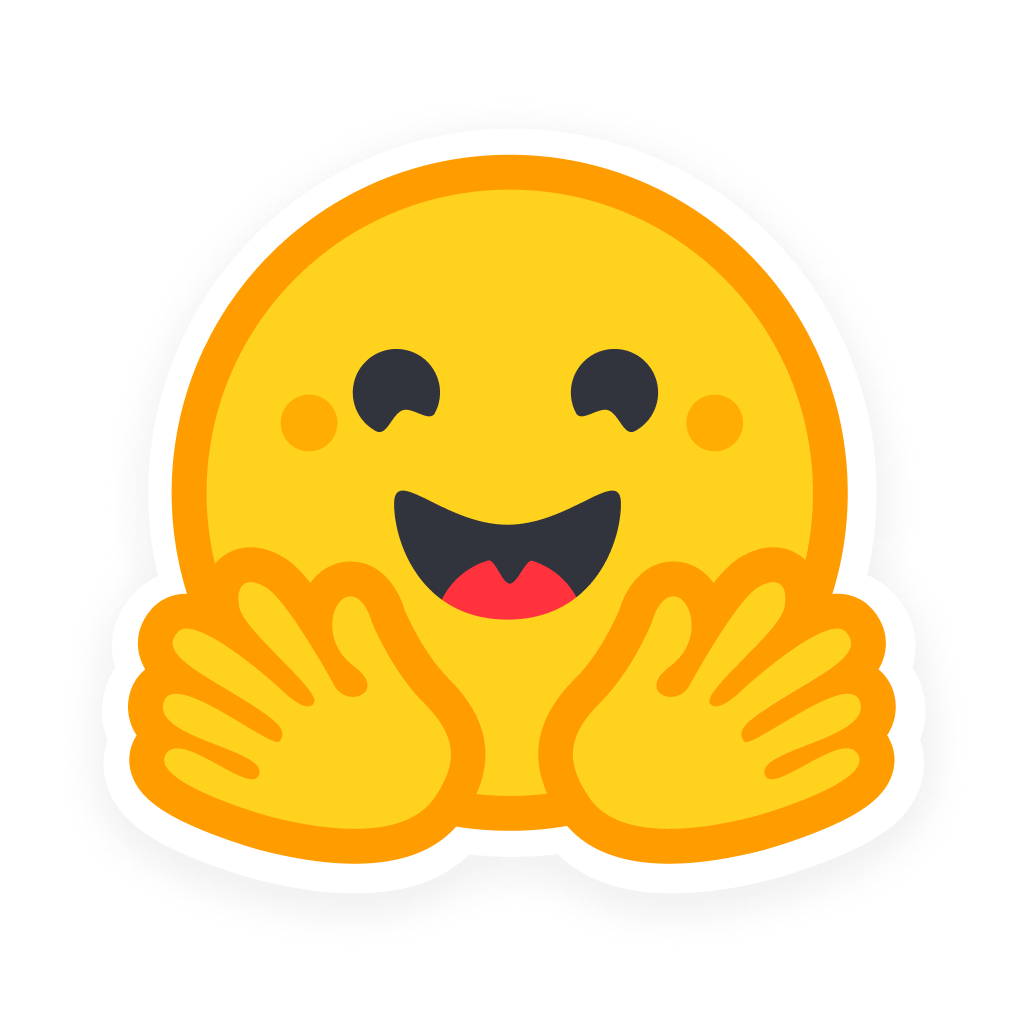} & {\footnotesize{\url{https://huggingface.co/climb-mao} \tiny{(CC BY 4.0)}}}
\end{tblr}
\end{abstract}

\section{Introduction} \label{introduction}

Curriculum Learning (CL) has emerged as a promising method to improve the cognitive plausibility of  \textbf{Small-Scale Language Models (SSLMs)} in the first BabyLM Challenge \cite{warstadt2023papers}, as a way to gradually introduce more complex linguistic phenomena into the model later in training in a manner that is similar to human language acquisition. Cognitively-inspired SSLMs are models trained on corpora that approximate the volume and nature of input that a first-language learner can expect to receive during language acquisition. These have been found to perform competitively against LLMs in English \cite{huebner-etal-2021-babyberta}.  CL strategies implemented in the BabyLM Challenge either specified a static measure of linguistic complexity, such as lexical frequency \cite{borazjanizadeh-2023-optimizing}, sorted datasets according to difficulty \cite{opper-etal-2023-effect}, or gradually increased vocabulary sizes \cite{edman-bylinina-2023-much}. While the majority of these strategies did not yield consistent improvements over  non-curriculum learning baselines \cite{warstadt2023papers}, linguistic theory suggests that children naturally focus on input that is neither too simple nor too difficult but at the right level of challenge for learning \cite{biberauer2019children, bosch_2023}. This is known as the “Goldilocks Effect”, which is a form of self-selecting curriculum learning that appears to naturally occur in first language (L1) acquisition. This raises the question of whether acquisition theories can provide insights into more effective curriculum learning strategies for SSLMs, and lead to more consistent benefits of CL strategies.  

Our work assesses whether language acquisition theories can provide us with better heuristics for good curriculum learning strategies to train SSLMs. We compare contrastive acquisition theories for their success when informing objective curriculum learning strategies on a standard architecture  \cite{martinez-etal-2023-climb}. We train SSLMs with three new objective curricula called \textsc{Growing, Inwards} and \textsc{MMM}, each replicating the developmental sequences of contemporary acquisition theories that first-language monolingual learners are theorised to follow in the earliest stages of acquisition cross-linguistically. In practice, these curricula modify the standard masked language modelling objective in  BabyBERTa-style models by varying the order and the sequence of masking using different tagsets to simulate different language acquisition theories. 

The acquisition models specify different cross-lingual and language-specific developmental sequences that learners appear to follow in first language acquisition, which has not been implemented or evaluated in the context of Deep Learning. The multilingual focus of the acquisition models is a goal strongly aligned with the spirit of the BabyLM Shared Task. We train SSLMs with these objective curricula for four typologically distant language families: Romance (French), Germanic (German), Japonic (Japanese) and Sino-Tibetan (Chinese). We introduce new age-ordered corpora of Child-Directed Speech (CDS) for these languages and select languages for pre-training based on the quantity of CDS that can be used to train SSLMs using similar volumes of data that learners can utilise in first language acquisition.  We evaluate these SSLMs on syntactic minimal pair datasets.  We find benefits of the cognitively-inspired objective curricula cross-linguistically, however different strategies lead to better performance for certain languages, particularly finer-grained language-specific versions of the \textsc{MMM} objective. Acquisition-inspired objective curricula can obtain comparable performance on minimal pair evaluation datasets to LLMs, despite requiring approximately 25X fewer parameters and 6,000X fewer words. 

\section{Background}  \label{background}

We survey Curriculum Learning (CL) strategies used in the \(1^{\text{st}}\) BabyLM Challenge \textit{Section} \ref{climb} and contrastive models of syntactic acquisition that are utilised to replicate cross-lingual developmental sequences for implementing more cognitively plausible pre-training in SSLMs in \textit{Section} \ref{acquisition}.

\subsection{Curriculum Learning Strategies for Pre-training on Developmentally Plausible Corpora} \label{climb}
 While some SSLMs that utilised CL strategies outperformed the official BabyLM baselines,  no CL strategies led to consistent or uniform improvements compared to stronger non-curriculum models.  Many submissions for the inaugural BabyLM Challenge utilised Curriculum Learning on a small-scale masked language model architecture trained on a 5 million (5M) word corpus called \textsc{BabyBERTa} \cite{huebner-etal-2021-babyberta}, based on a Transformer Language Model \textsc{RoBERTa} \cite{liu2019roberta} with \(15 \times\) fewer parameters, which displayed comparable grammatical capabilities to \textsc{RoBERTa}. In general, CL strategies, like using a pre-defined static difficulty assessment based on linguistic criteria like syntax dependency tree depth \cite{oba-etal-2023-babylm} or ranking sentences according to surprisal  \cite{chobey-etal-2023-training} or length \cite{debenedetto-2023-byte} or other measures of difficulty \cite{opper-etal-2023-effect},  showed little improvement over non-CL baselines. \citet{martinez-etal-2023-climb} introduce \textbf{Curriculum Learning for Infant-Inspired Model Building (CLIMB)}, which incorporates three CL strategies into BabyBERTa pre-training that each dynamically increase the difficulty of the language modelling task throughout training. CLIMB's \textbf{vocabulary curriculum} constrains the Transformer vocabulary in the initial stages of training by dynamically mask out vocabulary units over training. CLIMB's \textbf{data curriculum} varies the order of training instances based on infant-inspired expectations and the learning behaviour of the model, enabling dynamic sampling of training data according to a difficulty function. CLIMB's \textbf{objective curriculum} combines the masked language modelling task, used in RoBERTa \cite{liu2019roberta} and the BabyBERTa model \cite{huebner-etal-2021-babyberta}, with coarse-grained word class prediction to reinforce linguistic generalisation capabilities.  This provides functionality to change the objective function at specified discrete training steps. The objective curricula modifies the Masked Language Modelling (MLM) objective, which is the standard “denoising” objective for Pre-trained Language Models, like \textsc{RoBERTa} and \textsc{BabyBERTa}. Both models use a random token masking strategy, applying a fixed masking ratio \(\alpha\) to mask different contexts selected randomly with a probability \(P_i\).  \citet{martinez-etal-2023-climb} introduce two objective curricula defined using `curriculum units' of Universal Part of Speech (UPOS) tags. The first objective classifies \textsc{[mask]} to one of \textsc{[Verb, Noun, Other]}, while the second objective classifies \textsc{[mask]} to one of the 10 UPOS tags. CLIMB's objective curricula, following the submission guidelines of the 1st BabyLM Challenge, are performed using an unsupervised part-of-speech (POS) tagger.  They additionally tuned the vocabulary and model size of BabyBERTa, resulting in a model that outperformed the official baselines for the first BabyLM Challenge. CLIMB's curriculum learning strategies outperformed the official baseline but the accuracy of CL-strategies was comparable to the stronger BabyBERTa-style baseline introduced by the authors. We add new \textbf{cognitively-plausible objective curricula}, as an extension to the original CLIMB submission and CLIMB's improved  \textsc{BabyBERTa}-style as baselines.

\subsection{ Acquisition Models in Deep Learning: Three Models} \label{acquisition}
\begin{figure}[!ht]
        \centering
        \includegraphics[width=\linewidth]{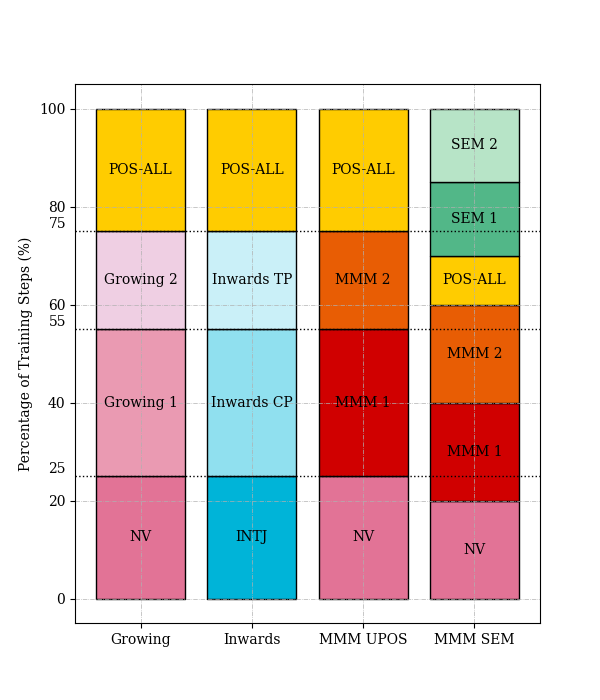}
        \caption{\textbf{ Acquisition-inspired Objective Curricula: }  We specify Objective Curricula \textsc{Growing, Inwards, MMM (UPOS), MMM (Semantic)} for three theories of acquisition (\textit{Section} \ref{acquisition}). The Progression of  
         Curriculum Units replicate the predicted developmental sequences by specifying curriculum units (defined in \textit{Table} \ref{tab:units}) defined over different pre-training stages, expressed as a percentage of training steps.}
        \label{fig:curricula}
\end{figure}

To assess whether using acquisition theories can be used to formulate better-performing CL strategies, we consider three recent language acquisition models that are amenable to Deep Learning implementation, as they specify developmental sequences that can be replicated as CL strategies in SSLMs. Based on careful linguistic analysis of universal and language-specific patterns in the utterances produced by learners cross-linguistically at different stages of acquisition, linguists have formalised strict (universal or non-language-specific) or weak (language-specific) orders of syntactic categories that are sequentially acquired. Since these acquisition models have been formulated based on linguistic analysis of multilingual acquisition data, we consider whether the CL strategies that precisely replicate these models can inform better-performing curriculum learning strategies cross-lingually. This leads us to train SSLMs with these objective curricula beyond English. As schematised in \textit{Figure} \ref{fig:curricula}, we can precisely replicate these developmental sequences as stages of SSLM pre-training, defined as proportions of training steps.  

We implement three contemporary cross-lingual models of syntactic acquisition: 

\begin{enumerate}
    \item \textbf{\textsc{Growing:}}   Bottom-up maturational approaches to language acquisition \cite{rizzi1993some,radford1990syntax}, including the “Growing Trees Hypothesis”\cite{Friedmann2021}, predicts that first language learners begin acquiring verbs and nouns (unit \textsc{NV} in \textit{Table} \ref{tab:units}). Learners subsequently progress to acquiring predicate information to form simple sentences; and finally, acquire discourse and complementiser information, allowing them to formulate complex sentences (e.g., with relative clauses). We can assume a tripartite model of bottom-up maturational development for implementation, with units \(\text{Growing } 1\) and \( \text{Growing } 2\) in \textit{Table} \ref{tab:units}.\footnote{There are differences in the number of stages predicted in bottom-up maturational approaches. Bottom-up approaches  \cite{rizzi1993some,radford1990syntax}  predict tripartite developmental sequence (a Verb Phrase, Tense Phrase and Complementiser Phrase), but Growing Trees involves bipartite stages (TP and VP is Stage 1, and Stage 2 involves acquiring the CP until QP to predict early acquisition of WH-questions).}
    \item \textbf{\textsc{Inwards}:} \citet{bosch_2023} introduces the predictions of a \textbf{generalised inward-growing maturational proposal (\textsc{Inwards})}, building on evidence from  \citet{HeimWiltschko2021} of early acquisition of “discourse”-material and interactional language (e.g. tags-questions). This predicts exactly the opposite order of acquisition of \textsc{Growing}. The stages of development begin with the early acquisition of complementisers used for illocutionary/discourse-related purposes (\textsc{Intj} and \textsc{Inwards- CP} in \textit{Table} \ref{tab:units}); followed by the acquisition of tense/event-related information (\textsc{Inwards-TP}); and finally, thematic information. 
    \item \textbf{\textsc{Neo-Emergent (MMM)}:} Neo-Emergentism predicts developmental stages in language acquisition that show increasing categorial granularity, taking a language-specific, or non-maturational, approach towards syntactic acquisition \cite{BiberauerRoberts2015}. The general universal prediction of one neo-emergent model called Maximise Minimal Means (MMM)  is that all learners, irrespective of the language being acquired, follow the same “coarse” stages in the acquisition of syntactic categories. They first learn to distinguish nouns and verbs (Unit \textsc{NV}), and then an “intermediate” set of categories (complementisers and event-related words),\footnote{In Chomskyan terminology, a vP-shell and a Complementiser Phrase (CP).} before finally learning tense/aspectual categories (units \(\text{MMM }  1 \) and \( \text{MMM } 2\) in \textit{Table} \ref{tab:units}). We implement this as a \textbf{universal “coarse” default curriculum strategy} that we implement as a default curriculum strategy (\textsc{MMM (upos)} in \textit{Figure} \ref{fig:curricula}). However, MMM also incorporates \textbf{language-specific differences in “finer-grained” curricula} where learners can acquire language-specific categories, leading to typological variation in the order of acquisition \cite{biberauer2019children, bosch_2023, bosch2024}, which we try to model in a CL strategy by specifying language-specific tagsets in \textsc{Sem 1, Sem 2} in \textit{Table} \ref{tab:units}.
\end{enumerate}

\begin{table}[!ht]
        \centering
        \small
        \begin{tabular}{|ll|} \hline 
        \textbf{Unit} &\textbf{POS Tags} \\ \hline \hline 
         NV & [NOUN, VERB]\\ \hline 
         Growing 1 & \(\textsc{NV} + \) [DET, ADJ, PRON,\\ 
          & PROPN, NUM, PRT]\\  \hline 
         Growing 2 & \(\text{growing}_1\) + [AUX, PART,\\ 
        & ADP, ADV]\\ \hline \hline 
         INTJ & [X, INTJ, SYM]\\  \hline 
        \textsc{Inwards CP} & \(\textsc{INTJ} + \) [PROPN,\\ 
        & CCONJ, SCONJ, SYM]\\  \hline 
        \textsc{Inwards TP} & \(\textsc{CP} + \) [NUM, PRT, AUX\\ 
        & PART, ADP, ADV]\\ \hline \hline  
         \textsc{MMM 1}& \(\textsc{NV} + \) [DET, CONJ, INTJ]\\  \hline 
         \textsc{MMM 2}& \textsc{MMM 1} + [ADJ, ADV, PRON,\\ 
         &   PROPN, NUM, PRT]\\ \hline \hline 
         \textsc{Sem 1}& \textsc{UPos} \(+ t_{\text{sem}} \in \) \textsc{[EVE,}\\
 &  TNS, ACT, ANA]\\  \hline 
         \textsc{Sem 2}& \textsc{sem1 + \(+ t_{\text{sem}} \in \)[ Log,}\\
 &\textsc{Com, Dem, Dis, Mod,}\\ 
         & \textsc{Ent, Nam, Tim]}\\ \hline 
        \end{tabular}
        \caption{Summary of Curriculum Units comprise Universal Part-of-Speech Tags and the Semantic Tags introduced by \citet{bjerva-etal-2016-semantic} used to define \textsc{Growing, Inwards \& MMM} objective curricula. The ordering of units for each acquisition-inspired curriculum is shown in \textit{Figure} \ref{fig:curricula}.}
        \label{tab:units}

\end{table}

Each stage of the \textsc{Growing, Inwards} and \textsc{MMM} models can be defined as a `curriculum unit' composed of POS tag sequences  listed in \textit{Table} \ref{tab:units}.\footnote{ The Chomskyan acquisition models used in this paper technically refer to syntactic projections, rather than part-of-speech tags.} To precisely replicate the developmental sequences of each acquisition model computationally, we will need to use a supervised tagger to specify curricula using strictly ordered sequences of POS tags. This is a cognitively motivated divergence from  \citet{martinez-etal-2023-climb}, who use an unsupervised tagger to define curricula. Using a supervised tagger is argued by   \citet{buttery2006computational} to enable computational modelling of a more cognitively plausible starting point for first language (L1) learners – based on a view of acquisition that is not fully emergent, nor completely nativist.\footnote{Note that \citet{buttery2006computational} uses a model  within a Combinatorial Categorial Grammar (CCG)-based formalism, which is also a “middle ground” between fully emergent acquisition models and a traditional biologically hardwired Universal Grammar assumed in traditional Chomskyan models like Principles and Parameters. }  For our purposes, it allows us to precisely replicate developmental sequences in SSLMs using curriculum learning.

\section{Dataset} \label{dataset}

\subsection{Training Corpora: \textsc{MAO-CHILDES}}
\begin{figure}[!ht]
    \centering
    \includegraphics[width=\linewidth]{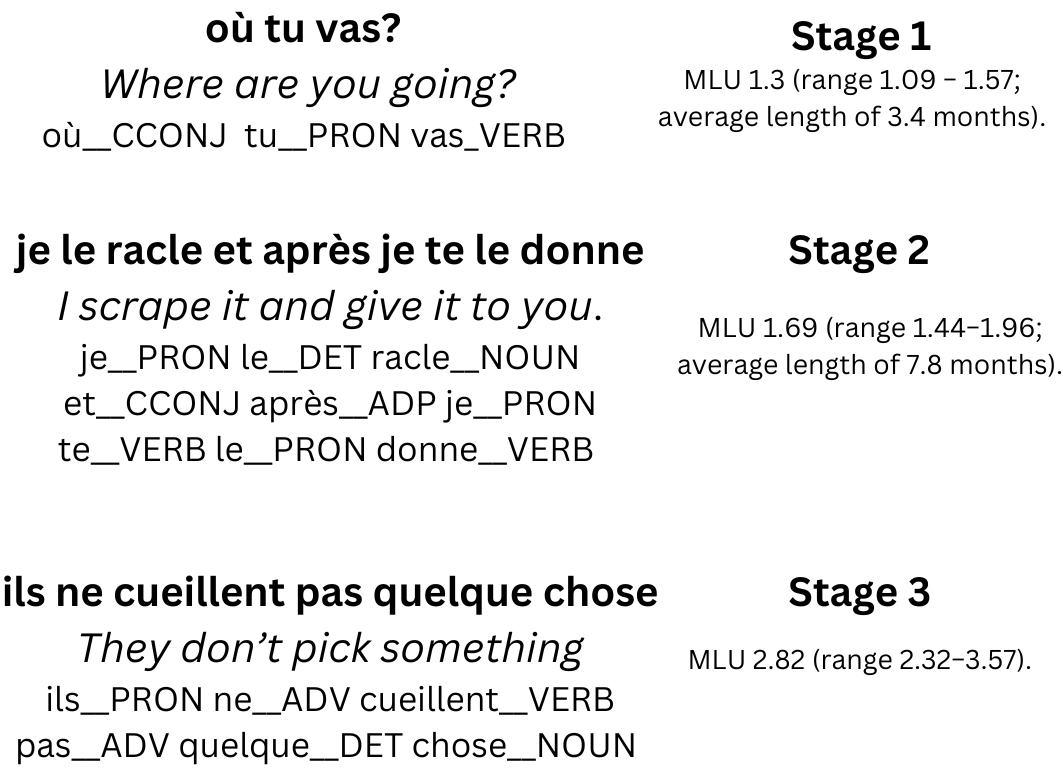}
        \caption{ A sample of Child-Directed Speech (CDS) from French\textbf{ \textsc{Mao-CHILDES}} that learners receive from caregivers at different stages of acquisition. Stages of acquisition are standardly defined in terms of mean lengths of utterances produced by learners.}
    \label{fig:fr-childes}
\end{figure}

We collect a training corpus of Age-ordered Child-Directed Speech (CDS) for four languages (French, German, Japanese and Chinese), in addition to the English Age-Ordered-CHILDES (AO-CHILDES) corpus \cite{huebner2021using} used in the BabyLM Challenge, to assess the benefits of the acquisition-inspired curricula beyond English compared to non-curriculum SSLMs. \textsc{Mao-CHILDES} is developed  from the Child Language  Data Exchange System (CHILDES) \cite{macwhinney2000childes}, which consists of in-home recordings of casual speech from caregivers to children and in-lab activities such as play, conversation and book reading directed towards first language learners for several languages.\footnote{Original data can be accessed here: \url{https://childes.talkbank.org/}}  We make our training corpus available on HuggingFace.\footnote{\url{https://huggingface.co/climb-mao}} The distribution of CHILDES data beyond English is a practical challenge for extending the BabyLM Challenge beyond English.  \textit{Table} \ref{tab:corpus-stats} shows the imbalance in quantities of CDS extracted from CHILDES, which is an artefact of a Western, Educated, Industrialised, Rich, and Democratic (WEIRD) bias in language acquisition research \cite{henrich2010weirdest}. A sample of CDS in the age-ordered corpora is shown in \textit{Figure} \ref{fig:fr-childes}, from different stages of language acquisition. Following \citet{huebner2021using}, utterances from children and child-directed speech (CDS) produced by caregivers, and other interlocuters, to children over the age of 6;0 are disregarded, leaving CDS produced by caregivers to children less than 6;0 which is sorted using the meta-data of the age of the learner in the CHILDES database.\footnote{The Script for Generating AO-CHILDES can be found here:\url{https://github.com/UIUCLearningLanguageLab/AOCHILDES}}   

\subsection{Evaluation Datasets}
To assess the success of three objective curricula (\textsc{Growing, Inwards} and \textsc{MMM}) that precisely replicate the predictions of the acquisition theories in \textit{Section} \ref{acquisition} on a standard SSLM architecture in a multilingual setting, we extend the evaluation pipeline of the BabyLM Challenge. This consists of syntactic evaluation datasets like BLiMP \cite{warstadt-etal-2020-blimp-benchmark} composed of minimal pairs of grammatical and ungrammatical sentences for language-specific syntactic phenomena. We use the following minimal pairs datasets to evaluate the objective curricula for the four languages in \textsc{Mao-CHILDES}:
\begin{enumerate}
    \item \textbf{CLAMS (French and German):} The Cross-Lingual Syntactic Evaluation of Word Prediction Models (CLAMS) \cite{mueller-etal-2020-cross}  generates minimal pair datasets which we use for French and German using Attribute-Varying Grammars. The dataset assesses grammaticality in Simple Agreement, VP coordination, and across “interveners” in S-V agreement (subject/object relative clause or across a Prepositional Phrase). 
    \item \textbf{JBLIMP (Japanese):} JBLIMP \cite{someya-oseki-2023-jblimp} is a minimal pairs dataset for targeted syntactic evaluation of Japanese. It consists of 331 minimal pairs of syntactic acceptability judgements curated from Japanese syntax articles in the \textit{Journal of East Asian Linguistics}.\footnote{The JBLiMP Minimal Pair dataset can be found here: \url{https://github.com/osekilab/JBLiMP/tree/main}} 
\item \textbf{SLING (Chinese):}  \textbf{SLING} \cite{song-etal-2022-sling} is a 38K minimal sentence pair dataset derived by applying syntactic and lexical transformations to Chinese Treebank 9.0,\footnote{The SLING Dataset can be found here: \url{https://huggingface.co/datasets/suchirsalhan/SLING}}  aiming to improve on the limitations of an earlier dataset called CLiMP  \cite{xiang-etal-2021-climp}, which had a lack of diversity in the vocabulary to generate minimal pair templates. 
\end{enumerate}

Due to the small size of the JBLIMP minimal pairs dataset, we follow \citet{someya-oseki-2023-jblimp}'s recommendation to compute accuracy using a SLOR score to mitigate the confounding effects of lexical frequencies and sentence lengths, which is defined as follows: 

\[SLOR(X) = \frac{\text{log} p_m (X) - \text{log} p_u (X)}{|X|}\]

where \(p_m(X)\) is the probability of a sentence for a Language Model and is the unigram probability of the sentence, estimated for each subword in the training corpus. Accuracy calculations for other languages follows dataset guidance to use unnormalised log-probabilities. 

\subsection{Universal POS Tagging}

 To define fine-grained objective curricula that perform masked language modelling with different subsets of syntactic and semantic tags for a specified proportion of training steps, we have to annotate child-directed speech corpora with Universal POS tags using an off-the-shelf SpaCy multilingual POS tagger.  The distribution of POS tags in \textsc{Mao-CHILDES} (Figure \ref{fig:5}) contains a high proportion of Nouns, whereas Verbs contribute a relatively low count. There are  orthographic issues in the CHILDES dataset for East Asian Languages, which are transcribed using Romanised characters (romaji) and a large proportion of English loan words in the Japanese portion of \textsc{Mao-CHILDES}, used in certain lexical domains, are incorrectly tagged automatically. These pre-processing inconsistencies were manually corrected. We also train a semantic tagger to specify language-specific curriculum strategies (see \textit{Appendix} \ref{MMM-SEM} for more detail). 

\begin{table*}[]
    \centering
    \begin{tabular}{|l|c|l|c|c|c|c|c|}
    \hline
         &\multicolumn{2}{|c|}{Model}& \textbf{English} & \textbf{Japanese} & \textbf{Chinese} & \textbf{French} & \textbf{German} \\ \hline
         Non-CL & \multicolumn{2}{|c|}{SSLM (\textsc{wiki})} & 64.60\% & 55.42\% & 48.01\% & 70.68\% & 59.63\% \\ \cline{2-8}
          & \multicolumn{2}{|c|}{\textsc{Mao-BabyBERTa}} & 75.48\% * & 61.21\% & 51.32\% & \textbf{80.00\%} & 68.78\% \\ \hline
         CL & \multicolumn{2}{|c|}{\textsc{Growing}} & 71.13\% & 79.30\% & 56.22\% & 76.21\% & 71.13\% \\ \cline{2-8}
          & \multicolumn{2}{|c|}{\textsc{Inwards}} & 71.05\% & 81.32\% & 54.26\% & 79.01\% & 69.34\% \\ \cline{2-8}
         & \textsc{MMM} & (\textsc{upos}) & 74.22\% & \textbf{87.31\%} & \textbf{58.79\%} & 75.93\% & \textbf{73.25\%} \\ 
          & & (\textsc{sem}) & \textbf{77.35\%} & & 55.01\% & & \\ \hline  
    \end{tabular}
    \caption{Evaluation of \textsc{Mao-BabyBERTa} (“vanilla” SSLM architecture without objective curricula) and the three Objective Curricula (\textsc{Growing}, \textsc{Inwards}, and \textsc{MMM}) on the following syntactic minimal pairs datasets: BLIMP (English), JBLIMP (Japanese), SLING (Chinese), CLAMS (French and German). Performance is compared to SSLM (\textsc{Wiki}). This is the same architecture trained on non-CDS training data. *This reports the performance of the best-performing “vanilla” model by \citet{martinez-etal-2023-climb} on the same architecture used to train our model. \textbf{Bolded} results indicate the highest accuracy of all the models.}
    \label{tab:results-overall}
\end{table*}

\section{Methodology}

\subsection{Model Architecture} \label{architecture}
Following \citet{martinez-etal-2023-climb}, we develop non-curriculum learning models. These models are scaled-down language models based on RoBERTa \cite{liu2019roberta}, with 8M parameters and trained on no more than 30M words \cite{huebner-etal-2021-babyberta}.  We use 8192 vocabulary items, which \citet{martinez-etal-2023-climb} find yields better overall performance compared to a larger vocabulary. Token unmasking is also removed, like BabyBERTa.  We use a small model architecture composed of eight layers. This follows \citet{martinez-etal-2023-climb}, who compare the role of model size (\(8,10, 12\) Transformer layers)  and vocabulary size (comparing \(|V | \in \{8192, 16 384 \}\)). An AdamW optimiser with linear scheduling is used \cite{loshchilov2017fixing}.  Each model is trained for 400,000 steps with 4 A100 GPUs. The hyperparameters used for the “vanilla” SSLMs are shown in \textit{Table} \ref{tab:climbhyperparameters}.  The models concatenate input sequences to capitalise on the available input length. 

\subsection{Baselines: LLMs and SSLM (\textsc{wiki})}
We use two families of models as baselines. First, we compare the performance of monolingual SSLMs to monolingual Large Language Models to assess the benefits of the BabyLM paradigm. For French, German and Chinese, we use RoBERTa-style monolingual LLMs.\footnote{The French RoBERTa model is available here: \url{https://huggingface.co/abhilash1910/french-roberta}. The German RoBERTa model is available here: \url{https://huggingface.co/uklfr/gottbert-base} } The Chinese RoBERTa model is  trained on around 30B words \cite{cui-etal-2020-revisiting}, which more than \(10^4\) times the training data we use to train our SSLMs in the Chinese portion of \textsc{Mao-CHILDES}.\footnote{The Chinese RoBERTa model is available here: \url{https://huggingface.co/hfl/chinese-roberta-wwm-ext-large}.} We include GPT-2 Baselines for Japanese, which are reported by \citet{someya-oseki-2023-jblimp}. This is because Japanese RoBERTa monolingual language models\footnote{Japanese RoBERTa models is available here:  \url{https://huggingface.co/rinna/japanese-roberta-base}} are not trained on data using Romaji orthography, which is used in the Japanese portion of \textsc{Mao-CHILDES} (\textit{Section} \ref{dataset}). Secondly, to assess the benefits of pre-training SSLMs on Child-Directed Speech, we train SSLMs using Wikipedia text (SSLM \textsc{wiki}), which is extracted to match the quantity of training data in \textsc{Mao-CHILDES} for each language. We keep the original hyperparameter settings used by \citet{huebner-etal-2021-babyberta}. 

\subsection{“Vanilla” SSLMs: MAO-BabyBERTa} \label{vanilla}

 We train a family of SSLMs, called Monolingual Age-Ordered BabyBERTa (\textsc{MAO-BabyBERTa}),  on language-specific training data from \textsc{Mao-CHILDES} using the model architecture described in \textit{Section} \ref{architecture} without any curriculum learning strategies.  Hyperparameters are tuned for English, and we use the same settings in MAO-BabyBERTa. 

\subsection{Implementing Acquisition-Inspired Objective Curricula: \textsc{Growing, Inwards \& MMM}} \label{cognitive}

To implement the acquisition-inspired strategies, we filter our age-ordered  \textsc{Mao-CHILDES} corpus for each language for expected utility in the acquisition process, according to the curriculum strategies of \textsc{Growing, Inwards} and \textsc{MMM} schematised in \textit{Figure} \ref{fig:curricula}. We then precisely implement the \textsc{Growing, Inwards, MMM} theories introduced in \textit{Section} \ref{acquisition}, using different curriculum units composed of POS tagsets  (\textit{Table} \ref{tab:units}) to define three objective curricula that replicate the developmental sequences of each acquisition model through the progressive ordering of POS units.  The logic for performing masked language modelling selectively for words annotated with a desired set of specified part of speech tags is implemented in \citet{martinez-etal-2023-climb}, which we extend. The objective curricula modify the masked language modelling (MLM) objective in a multi-task learning setup, so the acquisition-inspired objective is activated and optimised in parallel with MLM. We fix the model architecture to be identical to the “vanilla” SSLM architecture in \textit{Section} \ref{vanilla} to evaluate the benefits of each curriculum strategy. We modify CLIMB's objective curricula to implement the \textsc{Growing, Inwards} and \textsc{MMM} objective curricula by splitting \(400\)K training steps across four non-uniform intervals that are defined as a proportion of the SSLM's training steps, defined in \textit{Figure} \ref{fig:curricula}. This is meant to roughly simulate four developmental stages of an idealised monolingual learner until 6;0.  We then specify tagsets for each phase of the curricula that correspond to the acquisition theory. To illustrate this, the \textsc{Inwards} curriculum begins with a unit \textsc{INTJ}, which performs MLM for interjections and other interactional language, which are annotated with tags \textsc{Intj, X, Sym}. Then, we specify two further curriculum units \textsc{Inwards-CP} which performs MLM on complementiser-like words (e.g., \textsc{SConj}), and \textsc{Inwards-TP} which performs MLM on auxiliaries \textsc{Aux} and other tense/event-related words. At each stage of the curriculum, the objective curricula provide the vanilla SSLM model with a list of syntactic tags to use during training, taken from a pre-specified set of UPOS tags that lists all the tags used in the \textsc{Upos} tagged \textsc{Mao-CHILDES} training set.  If a tag is not used at the curriculum stage, its “ID” is set to zero so it is not a target for masked language modelling (MLM). During training, the number of part-of-speech tags that the model has to classify over are varied, according to the predictions of each acquisition model. The objective curricula end with a final curriculum unit, Pos-ALL, containing the entire Universal Part-of-Speech Tagset. The masking ratio is an important hyperparameter that impacts the pretraining of a Masked Language Model. A masking ratio of \(0.4\) is used for the tags specified at the curriculum stage. A \(0.15\) masking rate is used elsewhere if the tag is not specified at the curriculum stage.  For RoBERTa-based Language Models, a masking ratio of \(0.4\) performs better than \(0.15\) in downstream tasks  \cite{wettig-etal-2023-mask}.In addition to our “default” MMM strategy defined by Universal POS tags, \textsc{MMM (upos)}, we additionally introduce a \textbf{refined version of the MMM objective}, \textsc{MMM (sem)} for English and Chinese. This adds two additional stages to the non-language specific strategy to define a language-specific curricula that utilises semantic tags \cite{bjerva-etal-2016-semantic}, or \textit{sem}-tags, to model \textbf{language-specific acquisition strategies} (Section \ref{acquisition}). Detailed methods and results are discussed in \textit{Appendix} \ref{MMM-SEM}. Training times for each objective are summarised in \textit{Table} \ref{table:train}.

\section{Results}
The performance of objective curricula and cross-lingual SSLMs on minimal pairs datasets is summarised in \textit{Table} \ref{tab:results-overall}. \textbf{Fine-grained objective curricula demonstrate variable effectiveness compared to non-curriculum baselines}. While \textsc{MMM (upos)} shows general promise, average benefits of \textsc{MMM (upos)}, \textsc{Growing}, and \textsc{Inwards}, do not show statistically significant improvements on \textsc{Mao-BabyBERTa} cross-linguistically (\(p < 0.05\)). However, \textbf{the \textsc{MMM (SEM)} curriculum achieves a statistically significant performance improvement in both English and Chinese} (\(p < 0.05\)) when performing a paired t-test.  Instead, \textbf{statistically significant improvements are observed with acquisition-inspired CL strategies in specific languages across minimal pairs test sets}. \textsc{MMM (upos)} only achieves a statistically significant improvement in Japanese and Chinese. \textsc{Growing} leads to a statistically significant improvement in Japanese and Chinese, while  \textsc{Inwards} only has statistically significant improvements in Japanese. No curriculum strategy outperforms \textsc{Mao-BabyBERTa} in French, although \textsc{Inwards} almost reaches the same accuracy. German CL strategies only marginally outperform the non-CL baseline. In \textit{Figure} \ref{fig:english}, we compare these results with a broader range of models introduced by \citet{martinez-etal-2023-climb}, finding that the English \textsc{MMM (SEM)} curriculum marginally outperforms other curriculum learning strategies. See \textit{Appendix } \ref{significance} for details on how t-test statistics are computed. 

\begin{table}[ht]
\centering
\begin{tabular}{|l|c|c|}
\toprule \hline  
Language& LLM& SSLM (CL)\\ \hline
\midrule
English& \textbf{80.10} & 77.35(MMM \textsc{sem})\\ \hline 
Japanese& 77.95 & \textbf{87.31 (MMM)} \\ \hline 
Chinese& \textbf{83.41} &  58.79 (MMM) \\ \hline 
French& \textbf{83.00} &  79.01(Inwards) \\ \hline 
German& \textbf{92.16} &  73.25(MMM) \\
\bottomrule \hline
\end{tabular}
\caption{Comparison of Accuracy of LLMs and the Best Performing CL Strategy on Minimal Pairs Datasets. \textsc{Sem} represents Language-Specific strategies implemented for English and Chinese pre-training compared to the language-invariant \textsc{MMM (upos)} strategies.}
\label{tab:llm}
\end{table}

\section{Discussion}

Acquisition-inspired CL strategies represent a novel large-scale application of language acquisition theory in Deep Learning, aimed at improving the performance of SSLMs. Acquisition-inspired curricula guide SSLMs, which function as large statistical learners, to generalise over frequent linguistic categories—such as nouns and verbs—early in the training process and attend to language-specific features, such as the Germanic V2 word order. This suggests that \textbf{more fine-grained, language-specific curricula may have performance benefits over non-CL strategies in SSLMs}, which is supported by results showing the limited improvements of universal/maturational theories of acquisition that inform the \textsc{Growing} and \textsc{Inwards} strategies. Although both acquisition models predict universal curricula that should lead to consistent benefits cross-lingually, \textsc{Growing/Inwards} only improve performance in Chinese and Japanese, while performing comparably to non-curriculum (non-CL) baselines in French/German and worse than non-CL baselines in English. An additional benefit of using fine-grained language-specific curricula is that it enables SSLMs to learn more complex grammatical phenomena that may rely on semantics like anaphora. We notice notable improvements in ellipsis performance (\textit{Table} \ref{tab:blimp}) with the \textsc{MMM (Sem)} curriculum. Interestingly, in Chinese, the \textsc{MMM (sem)} curriculum marginally underperforms compared to \textsc{MMM (upos)} when handling anaphora and aspectual phenomena (\textit{Table} \ref{table:sling}), highlighting the need for further investigation into engineering optimal language-specific curriculum strategies that outperform non-CL strategies. This raises important avenues for future research. Careful analysis of developmental sequences beyond English to develop language-specific strategies similar to \textsc{MMM (upos/sem)} will be crucial. We encourage practitioners to curate larger corpora of child-directed speech (CDS) for training SSLMs in languages beyond English and to develop more  minimal pair datasets that have coverage beyond grammatical agreement in CLAMs to develop better-performing curriculum strategies for Romance and Germanic. Additionally, an important finding is that \textbf{acquisition-inspired CL strategies in Japanese significantly outperform GPT-2} (\textit{Table} \ref{tab:llm}). The improvements observed in Japanese control/raising phenomena (\textit{Table} \ref{table:jblimp}) suggest that the properties of CDS in Japanese may lead to more robust generalisations than LLMs.

\section{Conclusion}
This paper assesses whether fine-grained curriculum learning strategies based on acquisition theories can provide better heuristics for CL strategies for SSLM pre-training cross-lingually, introducing the \textsc{Mao-CHILDES} training corpus to train SSLMs for four typologically distant language families. Mixed results of the maturational \textsc{Growing} and \textsc{Inwards} acquisition theories in curriculum strategies and the implementation of the coarse/universal prediction of \textsc{MMM (upos)} suggest that there is no guaranteed performance benefit just by devising universal CL strategies based on acquisition theories for SSLMs in a multilingual setting. Training SSLMs using more fine-grained language-specific curricula that precisely replicate cutting-edge linguistic theories is effective for the \textsc{MMM (sem)} objective in English and Chinese and \textsc{MMM (upos)} in Japanese. Curriculum Learning can outperform non-curriculum SSLMs by specifying fine-grained language-specific curricula that precisely replicate language acquisition theories, highlighting how cognitively-inspired techniques can lead to better-performing data-efficient architectures in the spirit of the BabyLM Challenge.

\newpage 
 \section*{Acknowledgments}

 Many thanks to Andrew Caines for his comments, supervision and feedback on this paper. We thank N\'uria Bosch-Masip for her comments on the linguistic acquisition models implemented in this paper. We thank Mila Marcheva for her thoughts on cognitively-inspired modelling, which influenced the ideas in this paper. This paper reports on work supported by Cambridge University Press \& Assessment. It was performed using resources provided by the Cambridge Service for Data Driven Discovery (CSD3) operated by the University of Cambridge Research Computing Service, provided by Dell EMC and Intel using Tier-2 funding from the Engineering and Physical Sciences Research Council (capital grant EP/T022159/1), and DiRAC funding from the Science and Technology Facilities Council. Additionally, we thank the NVIDIA Corporation for the donation of the Titan X Pascal GPU used in this research. Richard Diehl Martinez is supported by the Gates Cambridge Trust (grant OPP1144 from the Bill \& Melinda Gates Foundation). Zébulon Goriely’s work is supported by The Cambridge Trust. 

\bibliography{anthology,custom}

\begin{thebibliography}{35}
\expandafter\ifx\csname natexlab\endcsname\relax\def\natexlab#1{#1}\fi

\bibitem[{Biberauer(2019)}]{biberauer2019children}
Theresa Biberauer. 2019.
\newblock \href {https://doi.org/10.1515/tl-2019-0013} {Children always go beyond the input: {T}he {M}aximise {M}inimal {M}eans perspective}.
\newblock \emph{Theoretical Linguistics}, 45(3-4):211--224.

\bibitem[{Biberauer and Roberts(2015)}]{BiberauerRoberts2015}
Theresa Biberauer and Ian Roberts. 2015.
\newblock \href {https://www.mmll.cam.ac.uk/files/copil_7_1_biberauerroberts.pdf} {Rethinking {F}ormal {H}ierarchies: {A} {P}roposed {U}nification}.
\newblock \emph{Cambridge Occasional Papers in Linguistics}, 7:1--31.

\bibitem[{Bjerva et~al.(2016)Bjerva, Plank, and Bos}]{bjerva-etal-2016-semantic}
Johannes Bjerva, Barbara Plank, and Johan Bos. 2016.
\newblock \href {https://aclanthology.org/C16-1333} {Semantic tagging with deep residual networks}.
\newblock In \emph{Proceedings of {COLING} 2016, the 26th International Conference on Computational Linguistics: Technical Papers}, pages 3531--3541, Osaka, Japan. The COLING 2016 Organizing Committee.

\bibitem[{Borazjanizadeh(2023)}]{borazjanizadeh-2023-optimizing}
Nasim Borazjanizadeh. 2023.
\newblock \href {https://doi.org/10.18653/v1/2023.conll-babylm.32} {Optimizing {GPT}-2 pretraining on {B}aby{LM} corpus with difficulty-based sentence reordering}.
\newblock In \emph{Proceedings of the BabyLM Challenge at the 27th Conference on Computational Natural Language Learning}, pages 356--365, Singapore. Association for Computational Linguistics.

\bibitem[{Bosch(2024)}]{bosch2024}
N{\'{u}}ria Bosch. 2024.
\newblock On another topic, how do acquisition orders vary? {T}he left periphery and topicalisation in bilinguals and monolinguals.
\newblock 1st year PhD report.

\bibitem[{Bosch(2023)}]{bosch_2023}
Núria Bosch. 2023.
\newblock \href {https://lingbuzz.net/lingbuzz/007418} {Emergent {S}yntax and {M}aturation: {A} {N}eo-{E}mergentist {A}pproach to {D}evelopment}.
\newblock \emph{MPhil Thesis, Department of Theoretical and Applied Linguistics, University of Cambridge}.

\bibitem[{Buttery(2006)}]{buttery2006computational}
Paula~J. Buttery. 2006.
\newblock \href {https://doi.org/10.48456/tr-675} {{Computational models for first language acquisition}}.
\newblock Technical Report UCAM-CL-TR-675, University of Cambridge, Computer Laboratory.

\bibitem[{Chobey et~al.(2023)Chobey, Smith, Wang, and Prasad}]{chobey-etal-2023-training}
Aryaman Chobey, Oliver Smith, Anzi Wang, and Grusha Prasad. 2023.
\newblock \href {https://doi.org/10.18653/v1/2023.conll-babylm.9} {Can training neural language models on a curriculum with developmentally plausible data improve alignment with human reading behavior?}
\newblock In \emph{Proceedings of the BabyLM Challenge at the 27th Conference on Computational Natural Language Learning}, pages 98--111, Singapore. Association for Computational Linguistics.

\bibitem[{Cui et~al.(2020)Cui, Che, Liu, Qin, Wang, and Hu}]{cui-etal-2020-revisiting}
Yiming Cui, Wanxiang Che, Ting Liu, Bing Qin, Shijin Wang, and Guoping Hu. 2020.
\newblock \href {https://doi.org/10.18653/v1/2020.findings-emnlp.58} {Revisiting pre-trained models for {C}hinese natural language processing}.
\newblock In \emph{Findings of the Association for Computational Linguistics: EMNLP 2020}, pages 657--668, Online. Association for Computational Linguistics.

\bibitem[{DeBenedetto(2023)}]{debenedetto-2023-byte}
Justin DeBenedetto. 2023.
\newblock \href {https://doi.org/10.18653/v1/2023.conll-babylm.17} {Byte-ranked curriculum learning for {B}aby{LM} strict-small shared task 2023}.
\newblock In \emph{Proceedings of the BabyLM Challenge at the 27th Conference on Computational Natural Language Learning}, pages 198--206, Singapore. Association for Computational Linguistics.

\bibitem[{Diehl~Martinez et~al.(2023)Diehl~Martinez, McGovern, Goriely, Davis, Caines, Buttery, and Beinborn}]{martinez-etal-2023-climb}
Richard Diehl~Martinez, Hope McGovern, Zebulon Goriely, Christopher Davis, Andrew Caines, Paula Buttery, and Lisa Beinborn. 2023.
\newblock \href {https://doi.org/10.18653/v1/2023.conll-babylm.10} {{CLIMB} {--} {C}urriculum {L}earning for {I}nfant-inspired {M}odel {B}uilding}.
\newblock In \emph{Proceedings of the BabyLM Challenge at the 27th Conference on Computational Natural Language Learning}, pages 112--127, Singapore. Association for Computational Linguistics.

\bibitem[{Edman and Bylinina(2023)}]{edman-bylinina-2023-much}
Lukas Edman and Lisa Bylinina. 2023.
\newblock \href {https://doi.org/10.18653/v1/2023.conll-babylm.8} {Too much information: Keeping training simple for {B}aby{LM}s}.
\newblock In \emph{Proceedings of the BabyLM Challenge at the 27th Conference on Computational Natural Language Learning}, pages 89--97, Singapore. Association for Computational Linguistics.

\bibitem[{Friedmann et~al.(2021)Friedmann, Belletti, and Rizzi}]{Friedmann2021}
Naama Friedmann, Adriana Belletti, and Luigi Rizzi. 2021.
\newblock \href {https://doi.org/10.5334/gjgl.131} {Growing trees: The acquisition of the left periphery}.
\newblock \emph{Glossa: a journal of general linguistics}, 6(1):131.

\bibitem[{Heim and Wiltschko(2021)}]{HeimWiltschko2021}
Jutta Heim and Martina Wiltschko. 2021.
\newblock Acquiring the form and function of interaction: a comparison of the acquisition of sentence-final particles and tag questions in the brown corpus.
\newblock Talk presented at LAGB Annual Meeting 2021 (online), 8 September.

\bibitem[{Henrich et~al.(2010)Henrich, Heine, and Norenzayan}]{henrich2010weirdest}
Joseph Henrich, Steven~J Heine, and Ara Norenzayan. 2010.
\newblock \href {https://www.cambridge.org/core/journals/behavioral-and-brain-sciences/article/abs/weirdest-people-inthe-world/BF84F7517D56AFF7B7EB58411A554C17} {The weirdest people in the world?}
\newblock \emph{Behavioral and {B}rain {S}ciences}, 33(2-3):61--83.

\bibitem[{Huebner et~al.(2021)Huebner, Sulem, Cynthia, and Roth}]{huebner-etal-2021-babyberta}
Philip~A. Huebner, Elior Sulem, Fisher Cynthia, and Dan Roth. 2021.
\newblock \href {https://doi.org/10.18653/v1/2021.conll-1.49} {{B}aby{BERT}a: Learning more grammar with small-scale child-directed language}.
\newblock In \emph{Proceedings of the 25th Conference on Computational Natural Language Learning}, pages 624--646, Online. Association for Computational Linguistics.

\bibitem[{Huebner and Willits(2021)}]{huebner2021using}
Philip~A Huebner and Jon~A Willits. 2021.
\newblock \href {https://doi.org/10.1016/bs.plm.2021.08.002} {Using lexical context to discover the noun category: Younger children have it easier}.
\newblock In \emph{Psychology of learning and motivation}, volume~75, pages 279--331. Elsevier.

\bibitem[{Li et~al.(2021)Li, Hou, Ye, Liang, and Sun}]{li-etal-2021-universal}
Wenxi Li, Yiyang Hou, Yajie Ye, Li~Liang, and Weiwei Sun. 2021.
\newblock \href {https://doi.org/10.18653/v1/2021.naacl-main.440} {Universal semantic tagging for {E}nglish and {M}andarin {C}hinese}.
\newblock In \emph{Proceedings of the 2021 Conference of the North American Chapter of the Association for Computational Linguistics: Human Language Technologies}, pages 5554--5566, Online. Association for Computational Linguistics.

\bibitem[{Liu et~al.(2019)Liu, Ott, Goyal, Du, Joshi, Chen, Levy, Lewis, Zettlemoyer, and Stoyanov}]{liu2019roberta}
Yinhan Liu, Myle Ott, Naman Goyal, Jingfei Du, Mandar Joshi, Danqi Chen, Omer Levy, Mike Lewis, Luke Zettlemoyer, and Veselin Stoyanov. 2019.
\newblock \href {https://arxiv.org/abs/1907.11692} {Ro{BERT}a: A {R}obustly {O}ptimized {BERT} {P}retraining {A}pproach}.
\newblock \emph{arXiv preprint arXiv:1907.11692}.

\bibitem[{Loshchilov et~al.(2017)Loshchilov, Hutter et~al.}]{loshchilov2017fixing}
Ilya Loshchilov, Frank Hutter, et~al. 2017.
\newblock \href {https://openreview.net/forum?id=rk6qdGgCZ} {Fixing weight decay regularization in adam}.
\newblock \emph{arXiv preprint arXiv:1711.05101}, 5.

\bibitem[{MacWhinney(2000)}]{macwhinney2000childes}
Brian MacWhinney. 2000.
\newblock \href {https://doi.org/10.4324/9781315805641} {\emph{The {CHILDES} {P}roject: The {D}atabase}}, volume~2.
\newblock Psychology Press.

\bibitem[{Marcus et~al.(1993)Marcus, Marcinkiewicz, and Santorini}]{marcus}
Mitchell~P. Marcus, Mary~Ann Marcinkiewicz, and Beatrice Santorini. 1993.
\newblock \href {https://aclanthology.org/J93-2004.pdf} {Building a {L}arge {A}nnotated {C}orpus of {E}nglish: {T}he {P}enn {T}reebank}.
\newblock \emph{Comput. Linguist.}, 19(2):313–330.

\bibitem[{Mueller et~al.(2020)Mueller, Nicolai, Petrou-Zeniou, Talmina, and Linzen}]{mueller-etal-2020-cross}
Aaron Mueller, Garrett Nicolai, Panayiota Petrou-Zeniou, Natalia Talmina, and Tal Linzen. 2020.
\newblock \href {https://doi.org/10.18653/v1/2020.acl-main.490} {Cross-linguistic syntactic evaluation of word prediction models}.
\newblock In \emph{Proceedings of the 58th Annual Meeting of the Association for Computational Linguistics}, pages 5523--5539, Online. Association for Computational Linguistics.

\bibitem[{Oba et~al.(2023)Oba, Haga, Fukatsu, and Oseki}]{oba-etal-2023-babylm}
Miyu Oba, Akari Haga, Akiyo Fukatsu, and Yohei Oseki. 2023.
\newblock \href {https://doi.org/10.18653/v1/2023.conll-babylm.25} {{B}aby{LM} challenge: Curriculum learning based on sentence complexity approximating language acquisition}.
\newblock In \emph{Proceedings of the BabyLM Challenge at the 27th Conference on Computational Natural Language Learning}, pages 290--297, Singapore. Association for Computational Linguistics.

\bibitem[{Opper et~al.(2023)Opper, Morrison, and Siddharth}]{opper-etal-2023-effect}
Mattia Opper, J.~Morrison, and N.~Siddharth. 2023.
\newblock \href {https://doi.org/10.18653/v1/2023.conll-babylm.31} {On the effect of curriculum learning with developmental data for grammar acquisition}.
\newblock In \emph{Proceedings of the BabyLM Challenge at the 27th Conference on Computational Natural Language Learning}, pages 346--355, Singapore. Association for Computational Linguistics.

\bibitem[{Radford(1990)}]{radford1990syntax}
Andrew Radford. 1990.
\newblock \href {https://doi.org/10.1207/s15327817la0103_1} {The {S}yntax of {N}ominal {A}rguments in {E}arly {C}hild {E}nglish}.
\newblock \emph{Language Acquisition}, 1(3):195--223.

\bibitem[{Rizzi(1993)}]{rizzi1993some}
Luigi Rizzi. 1993.
\newblock \href {https://doi.org/10.1207/s15327817la0304_2} {Some {N}otes on {L}inguistic {T}heory and {L}anguage {D}evelopment: The case of root infinitives}.
\newblock \emph{Language {A}cquisition}, 3(4):371--393.

\bibitem[{Salhan(2023)}]{salhan_2023}
Suchir~A. Salhan. 2023.
\newblock \href {https://www.mmll.cam.ac.uk/files/v15_salhan_2.pdf} {On the potential for ‘{M}aximising {M}inimal {M}eans’ in {T}ransformer {L}anguage {M}odels: {A} {D}ynamical {S}ystems {T}heory {P}erspective}.
\newblock \emph{Cambridge {O}ccasional {P}apers in {L}inguistics}, page 55–110.

\bibitem[{Someya and Oseki(2023)}]{someya-oseki-2023-jblimp}
Taiga Someya and Yohei Oseki. 2023.
\newblock \href {https://doi.org/10.18653/v1/2023.findings-eacl.117} {{JBL}i{MP}: {J}apanese benchmark of linguistic minimal pairs}.
\newblock In \emph{Findings of the Association for Computational Linguistics: EACL 2023}, pages 1581--1594, Dubrovnik, Croatia. Association for Computational Linguistics.

\bibitem[{Song et~al.(2022)Song, Krishna, Bhatt, and Iyyer}]{song-etal-2022-sling}
Yixiao Song, Kalpesh Krishna, Rajesh Bhatt, and Mohit Iyyer. 2022.
\newblock \href {https://doi.org/10.18653/v1/2022.emnlp-main.305} {{SLING}: {S}ino linguistic evaluation of large language models}.
\newblock In \emph{Proceedings of the 2022 Conference on Empirical Methods in Natural Language Processing}, pages 4606--4634, Abu Dhabi, United Arab Emirates. Association for Computational Linguistics.

\bibitem[{Warstadt et~al.(2023)Warstadt, Choshen, Mueller, Williams, Wilcox, and Zhuang}]{warstadt2023papers}
Alex Warstadt, Leshem Choshen, Aaron Mueller, Adina Williams, Ethan Wilcox, and Chengxu Zhuang. 2023.
\newblock \href {http://arxiv.org/abs/2301.11796} {Call for papers -- {T}he {B}aby{LM} {C}hallenge: {S}ample-efficient pretraining on a developmentally plausible corpus}.

\bibitem[{Warstadt et~al.(2020)Warstadt, Parrish, Liu, Mohananey, Peng, Wang, and Bowman}]{warstadt-etal-2020-blimp-benchmark}
Alex Warstadt, Alicia Parrish, Haokun Liu, Anhad Mohananey, Wei Peng, Sheng-Fu Wang, and Samuel~R. Bowman. 2020.
\newblock \href {https://doi.org/10.1162/tacl_a_00321} {{BL}i{MP}: The benchmark of linguistic minimal pairs for {E}nglish}.
\newblock \emph{Transactions of the Association for Computational Linguistics}, 8:377--392.

\bibitem[{Wettig et~al.(2023)Wettig, Gao, Zhong, and Chen}]{wettig-etal-2023-mask}
Alexander Wettig, Tianyu Gao, Zexuan Zhong, and Danqi Chen. 2023.
\newblock \href {https://doi.org/10.18653/v1/2023.eacl-main.217} {Should you mask 15{\%} in masked language modeling?}
\newblock In \emph{Proceedings of the 17th Conference of the European Chapter of the Association for Computational Linguistics}, pages 2985--3000, Dubrovnik, Croatia. Association for Computational Linguistics.

\bibitem[{Xiang et~al.(2021)Xiang, Yang, Li, Warstadt, and Kann}]{xiang-etal-2021-climp}
Beilei Xiang, Changbing Yang, Yu~Li, Alex Warstadt, and Katharina Kann. 2021.
\newblock \href {https://doi.org/10.18653/v1/2021.eacl-main.242} {{CL}i{MP}: A benchmark for {C}hinese language model evaluation}.
\newblock In \emph{Proceedings of the 16th Conference of the European Chapter of the Association for Computational Linguistics: Main Volume}, pages 2784--2790, Online. Association for Computational Linguistics.

\bibitem[{Xue et~al.(2005)Xue, Xia, Chiou, and Palmer}]{xue2005penn}
Naiwen Xue, Fei Xia, Fu-Dong Chiou, and Marta Palmer. 2005.
\newblock \href {https://doi.org/10.1017/S135132490400364X} {The {P}enn {C}hinese {T}reebank: Phrase structure annotation of a large corpus}.
\newblock \emph{Natural language engineering}, 11(2):207--238.

\end{thebibliography}
\bibliographystyle{acl_natbib}

\newpage 
\appendix

\section{\textsc{MMM (Sem)}: Specifying Language-Specific Curricula using Semantic Tags} \label{MMM-SEM}
As a first step towards modelling language-specific curricula using curriculum learning, we use Universal Semantic Tagging (\textit{sem-tagging}) \cite{bjerva-etal-2016-semantic}. The set of semantic tags can differ cross-lingually. In Chinese, \citet{li-etal-2021-universal} specifies a language-specific semantic tagset, adding and removing tags based on Chinese's semantic and syntactic properties. The fine-grained curriculum in an SSLM set-up aims to circumvent known problems of shortcut learning in LLMs  that prevent Transformer-based models from exhibiting robust structural generalisation capabilities that humans exhibit in acquisition  \cite{salhan_2023}. 

We perform \textit{sem}-tagging to annotate the BabyLM corpus for English and the Chinese corpus in \textsc{Mao-CHILDES} with  a set of language-neutral tags (sem-tags).  For English, we only perform \textit{sem}-tagging for the Adult Directed Speech datasets in the BabyLM Challenge dataset: the BNC,  Project Gutenberg , OpenSubtitles, QCRI, Wikipedia and Switchboard corpora. This allows us to modify our UPOS curricula for English to specify a more complex curricula to simulate later stages of language acquisition. The first stage of the new MMM curriculum  using semantic tags includes tags related to event, \texttt{EVE},  tense, \texttt{TNS}, and modality \texttt{MOD}. These are typically learnt later during acquisition, as part of complex tense sequences of auxiliaries and modal verbs \cite{BiberauerRoberts2015}, and allow us to define a \textbf{language-specific} sem-tag objective.  For Chinese, we sem-tag a  corpus of Wikipedia text that contains the same amount of text as the age-ordered CHILDES corpora introduced in \textit{Section} \ref{dataset}.

\subsection{Semantic Tagger Accuracy}

A multi-objective POS and \textit{sem}-tagger is trained,  using a Bidirectional LSTM (BiLSTM) with a Conditional Random Field (CRF) inference layer to train a multi-objective semantic and UPOS tagger for English and Chinese. This is trained on 1100 \textit{sem}-tagged sentences from the Wall Street Journal (WSJ) section of the Penn Treebank \cite{marcus}  and a 1000 \textit{sem}-tagged sentences from  Chinese TreeBank \cite{xue2005penn} annotated by \citet{li-etal-2021-universal}. The tagger has \(91.4 \%\) accuracy for Chinese and \(94.6 \%\) accuracy for English. 

\newpage 
\section{Training}
\begin{table}[!ht]
    \centering
    \caption{Hyperparameter Settings for CLIMB's “vanilla” and curriculum models and MAO-BabyBERTa (CDS)}
    \begin{tabular}{|c|c|} \hline 
         Layers&  \(8\) \\ \hline 
         Heads& 
    \(8\) \\ \hline 
 Hidden&\(256\) \\ \hline 
 \( |V|\)&\(8,192\) \\ \hline 
 Layer Norm EPS &\(1 \times 10^{-5}\) \\ \hline 
 Learning Rate &\(0.001\) \\ \hline 
 Optimizer&AdamW \\ \hline 
 Scheduler Type&Linear \\ \hline 
 Max Steps&\(400,000\) \\ \hline 
 Warm-up Steps &\(100,000\) \\ \hline\end{tabular}

    \label{tab:climbhyperparameters}
\end{table}

\begin{table}[h!]
\small
\centering
\begin{tabular}{|l|l|l|}
\hline \hline 
\textbf{Type} & \textbf{Model} & \textbf{Training Time} \\ \hline \hline
 \textsc{Mao-CLIMB}& \textsc{Growing}&11h 51m\\ \hline 
 & \textsc{Inwards}&11h 51m\\ \hline 
 & \textsc{MMM (upos)}&11h 46m\\ \hline 
 & \textsc{MMM (sem)}&25h 3m\\\hline 
Vanilla Models & CLIMB-small-raw & 12h \\ \hline \hline
\end{tabular}
\caption{Compute required to train our models. We report the model with the shortest and longest runtime for each experiment type. Each model is trained for 400,000 steps with 4 A100 GPUs.}
\label{table:train}
\end{table}

\section{Statistical Significance \& Detailed Results} \label{significance}
The statistical significance of the three curriculum strategies, \textsc{Growing, Inwards \& MMM} is calculated by performing t-tests on the detailed results in \textit{Tables} \ref{tab:blimp}, \ref{table:sling}, \ref{table:jblimp}, \ref{table:clams}. For each curriculum (\textsc{Growing}, \textsc{Inwards}, \textsc{MMM} (UPOS), \textsc{MMM} (SEM)), we calculate the paired differences in accuracy with the Vanilla model for all the test sets in the minimal pairs evaluation dataset. We perform paired t-tests for the non-CL baseline (\textsc{Mao-BabyBERTa}) and the accuracy of the respective curriculum for each curriculum strategy for each language, concluding that the curriculum-based model significantly outperforms the Vanilla/MaoBabyBERTa model if the \textit{p}-value is below our significance level \(\alpha = 0.05\).  The detailed results, below, support the findings of \citet{huebner-etal-2021-babyberta} cross-linguistically of the benefits of using less training data and paying careful attention to training artefacts and the domain of training corpora, as using CDS to train SSLMs (with/without objective curricula)  outperforms \textsc{SSLM (wiki)}.  

\newpage 

\begin{figure*}[!ht]
    \centering
        \caption{\textbf{Comparision of BLiMP Performance of English SSLMs with CLIMB curricula and \textsc{Growing, Inwards, MMM (UPOS), MMM (SEM)} (\textit{Section} \ref{cognitive})}  We report introduced by \citet{warstadt2023papers} for T5-base and OPT-125m models. We include the improved BabyBERTa baseline implemented in \citet{martinez-etal-2023-climb}, which beat the baseline used in the \( 1^{\text{st}}\) BabyLM Shared Task. We report BLiMP performance of different CLIMB \texttt{small-raw} models (also used in the standard architecture of \textsc{Mao-BabyBERTa} used with the three objective curricula) for the best performing dynamic curriculum learning strategies implemented in \citet{martinez-etal-2023-climb}. This includes CLIMB's \textbf{Data Curriculum} (Log Pacing with Source Difficulty), \textbf{Vocabulary Curriculum} (Log Pacing with Token ID Difficulty),  two \textbf{Objective Curricula strategies} (\textsc{MLM + All} uses a multitask objective of masked language modelling and objective curricula specified by 10 tags throughout all training steps, \textsc{MLM + NV} uses three tags throughout training), and the best performing \textbf{Combination Model}  (Token ID Vocabulary Curricula, Random + model ppx Data Curricula,  Multitask Objective Curricula).}
    \includegraphics[width=\linewidth]{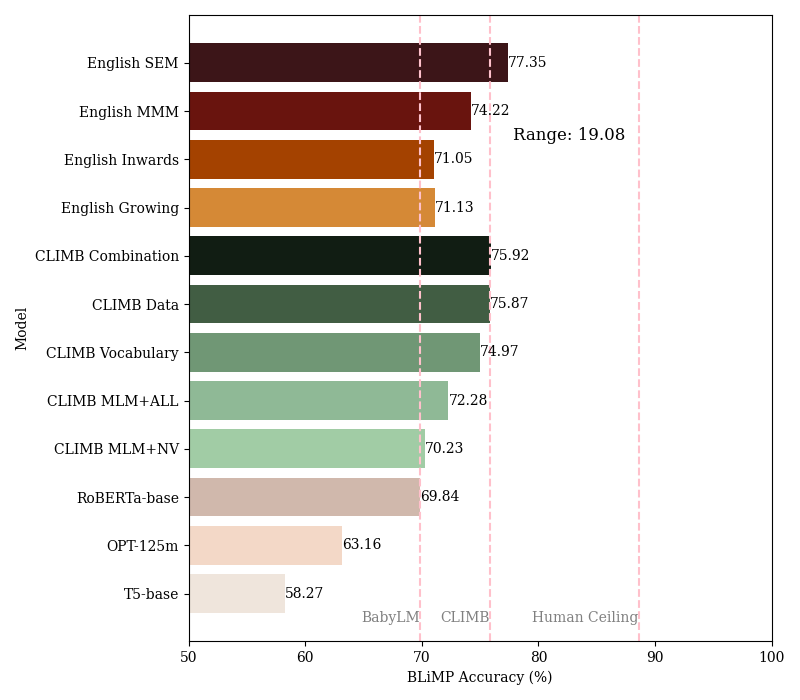}
    \label{fig:english}
\end{figure*}

\newpage 

\begin{table*}[!ht]
\small
    \centering
    \caption{Corpus Statistics for the \textbf{Child-Directed Speech (CDS)} files extracted from CHILDES for 24 languages, which are used  to select four languages for training.  The \textsc{Mao-CHILDES} corpus is selected based on the frequency of CDS, along additional considerations of evaluation. }
    \label{tab:corpus-stats}
    \begin{tabular}{|@{}l|l|l|l|l|l|l@{}|}
        \toprule \hline \hline
        lang   &  Samples & \(|V|\) & Tokens & Sentence Length  \(\mu\) & Children & Utterances\\ \hline\hline
        \textbf{Chinese}    & 857,792          & 518,172         & 850,510      & 258.28               & 949      & 3,293      \\
 \textbf{German}& 582192& 516,147& 867,704& 107.05& 65&8105\\\hline \hline 
        \textbf{Japanese}   & 537,164          & 280,807         & 528,930      & 38.67                & 122      & 13,678     \\\hline \hline 
        Indonesian & 537,235          & 286,448         & 521,759      & 202.31               & 9        & 2,579      \\\hline \hline 
        \textbf{French}     & 488,094          & 284,381         & 469,258      & 175.69               & 204      & 2,671      \\ \hline 
        \textbf{Spanish}    & 332,903          & 211,559         & 331,009      & 167.85               & 291      & 1,972      \\ \hline 
        \textbf{Dutch}      & 261,786          & 160,520         & 259,263      & 97.50                & 96       & 2,659      \\ \hline 
        \textbf{Portuguese} & 100,512          & 59,205          & 98,620       & 39.72                & 195      & 2,483      \\ \hline 
        Polish     & 82,977           & 71,072          & 82,940       & 43.04                & 14       & 1,927      \\ \hline 
        Swedish    & 80,936           & 53,719          & 79,739       & 49.34                & 6        & 1,616      \\ \hline 
        Norwegian  & 55,262           & 31,310          & 40,215       & 32.62                & 6        & 1,233      \\ \hline 
        Catalan    & 54,518           & 37,250          & 53,157       & 29.73                & 7        & 1,788      \\ \hline 
        Romanian   & 33,130           & 20,700          & 32,986       & 16.58                & 6        & 1,990      \\ \hline 
        Croatian   & 51,948           & 36,922          & 51,809       & 27.33                & 3        & 1,896      \\ \hline 
        Czech      & 45,122           & 33,185          & 44,117       & 27.15                & 6        & 1,625      \\ \hline 
        Danish     & 44,909           & 25,039          & 44,909       & 24.94                & 2        & 1,801      \\ \hline 
        Bulgarian  & 31,715           & 21,435          & 31,714       & 32.76                & 1        & 968        \\ \hline 
        Afrikaans  & 22,021           & 18,475          & 21,984       & 18.68                & 52       & 1,177      \\ \hline 
        Irish      & 18,973           & 13,598          & 18,869       & 9.82                 & 5        & 1,921      \\ \hline 
        Russian    & 7,008            & 5,963           & 7,007        & 4.42                 & 2        & 1,585      \\ \hline 
        Icelandic  & 47,945           & 27,775          & 46,516       & 11.36                & 1        & 4,094      \\ \hline 
        Slovenian  & 1,384            & 1,243           & 1,382        & 10.39                & 1& 133        \\ \hline
        Thai       & 38,550           & 27,084          & 38,329       & 100.34               & 18       & 382        \\ \hline \hline
    \end{tabular}
    
\end{table*}

\begin{figure*}[!ht]
    \centering
    \caption{Distribution of Silver Tags across all languages in the \textsc{Mao-CHILDES} corpus, annotated using a SpaCy Multilingual UPOS Tagger}
    \includegraphics[width=\linewidth]{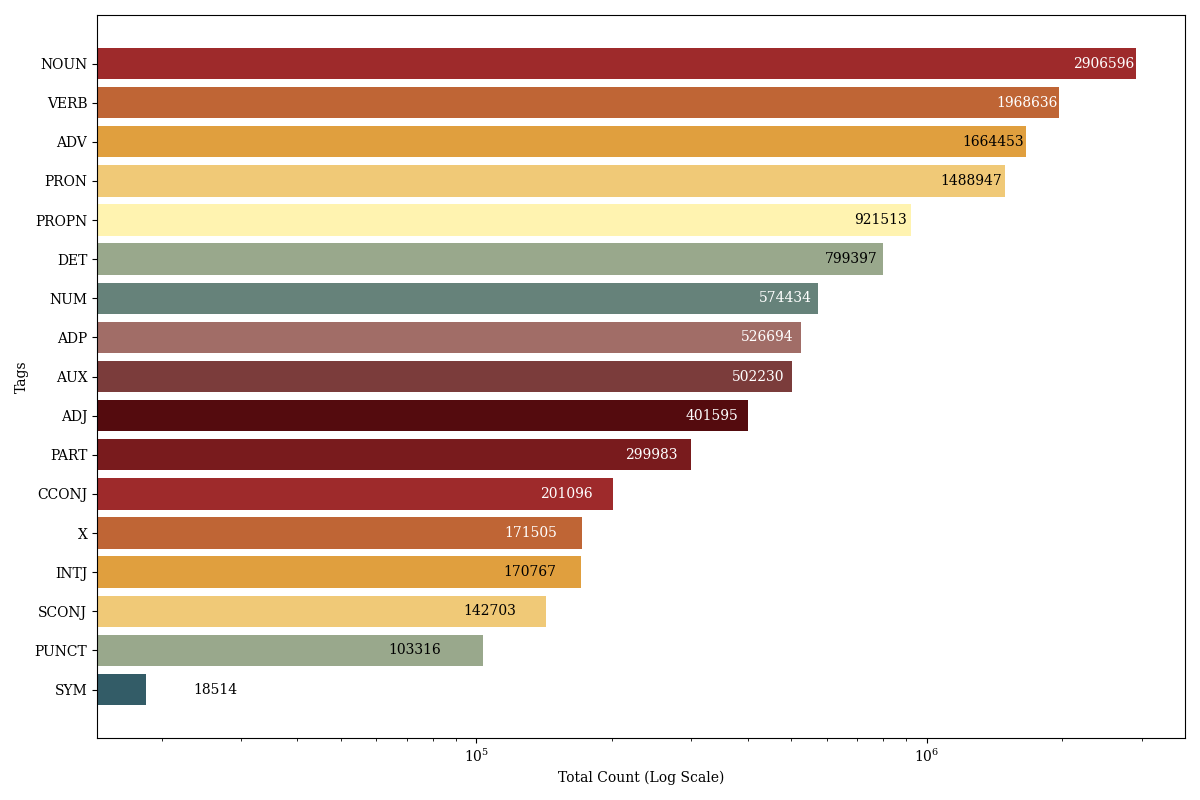}
    
    \label{fig:5}
\end{figure*}

\newpage 

\begin{table*}[h]
\small
\centering
\caption{\textbf{(English) }Evaluation of BabyBERTa model with four Cognitively-Plausible Curriculum Learning Strategies on BLIMP. English \textsc{Growing} based on “Growing Trees” \cite{Friedmann2021}, \textsc{Inwards} based on “Inward Maturation” \cite{HeimWiltschko2021} and \textsc{MMM (UPOS)} and the language-specific \textit{sem}-tag \textsc{MMM (SEM)} curricula based on \citet{BiberauerRoberts2015}. }
\label{tab:blimp}
\begin{tabular}{|@{}l|c|c|c@{}|l|}
\toprule \hline  \hline 
Grammatical Phenomenon& \textbf{Growing}& \textbf{Inwards}& \textbf{MMM} (UPOS) &\textbf{MMM} (SEM)\\ \midrule \hline  \hline 
Anaphor & \(\textbf{96.22\%}\)      & \(84.67\%\)             & \(81.13\%\)           &\(\textbf{90.89\%}\)\\ \hline 
Arg Str& \(79.13\%\)               & \(79.86\%\)             & \(84.79\%\)  &\(\textbf{85.99\%}\)\\ \hline 
Binding& \(46.47\%\)               & \(71.75\%\)             & \(\textbf{83.42\%}\)  &\(77.76\%\)\\ \hline 
Control-Raising& \(77.03\%\)               & \(73.82\%\)             & \(\textbf{88.02\%}\)  &\(82.10\%\)\\ \hline 
Det-N Agreement& \(65.49\%\)               & \(65.19\%\)             & \(\textbf{84.38\%}\)  &\(79.31\%\)\\ \hline 
Ellipsis& \(58.24\%\)      & \(53.26\%\)             & \(42.77\%\)           &\(\textbf{70.94\%}\)\\ \hline 
Filler Gap& \(80.70\%\)               & \(\textbf{88.47\%}\)    & \(85.60\%\)           &\(73.11\%\)\\ \hline 
Irregular& \(\textbf{76.34\%}\)      & \(44.85\%\)             & \(54.42\%\)           &\(74.91\%\)\\ \hline 
Island& \(69.53\%\)               & \(62.87\%\)             & \(\textbf{96.62\%}\)  &\(68.64\%\)\\ \hline 
NPI& \(69.21\%\)               & \(76.02\%\)             & \(\textbf{83.42\%}\)  &\(74.13\%\)\\ \hline 
Quantifiers& \(44.54\%\)               & \(\textbf{84.79\%}\)    & \(58.43\%\)           &\(71.86\%\)\\ \hline 
Subject-Verb& \(65.98\%\)               & \(64.89\%\)             & \(\textbf{68.37\%}\)  &\(\textbf{79.03\%}\)\\ \midrule \hline \hline 
\textbf{Average Accuracy}      & \(71.13\%\)               & \(71.05\%\)             & \(74.22\%\)& \(\textbf{77.35\%}\)\\ \bottomrule \hline \hline 
\end{tabular}
\end{table*}

\begin{table*}[htbp]
    \small
    \centering
        
    \caption{\textbf{(Chinese) } Comparison of accuracy of Chinese \textsc{Mao-BabyBERTa} (“vanilla”) and \textsc{Growing, Inwards, MMM (upos), MMM (sem)} objective curricula compared to a Chinese RoBERTa LLM baseline on the SLING minimal pairs dataset \cite{song-etal-2022-sling} }
    \begin{tabular}{|l| l |c | c | c | c |c |c|}
        \toprule  \hline \hline
        \textbf{Category} & \textbf{Subcategory} & \textbf{Vanilla}& \textbf{LLM} & \textbf{Growing} & \textbf{Inwards} & \textbf{MMM }& \textbf{MMM }\\
 & & & & & & \textbf{ (UPOS)}&\textbf{(SEM)}\\ \hline \hline 
        \midrule
        RelativeClause & rc\_resumptive\_pronoun & 50.50& 60.30& 50.50& 49.50& 53.10& 50.70
\\
        RelativeClause & rc\_resumptive\_noun & 48.00& 27.60& 48.90& 47.80& 58.00& 48.50
\\
        Anaphor & baseline\_female & 86.70& 75.60& 86.30& 83.90& 36.90& 85.60
\\
        Anaphor & pp\_female & 70.50& 71.80& 70.80& 67.50& 41.80& 69.80
\\
        Anaphor & baseline\_male & 12.50& 38.50& 45.20& 45.30& 81.90& 45.20
\\
        Anaphor & Plural & 51.98& 97.95& 53.10& 51.20& 52.33& 52.10
\\
        Anaphor & self\_male & 14.30& 92.60& 47.80& 46.10& 81.40& 46.90
\\
        Anaphor & pp\_male & 28.00& 77.60& 49.50& 48.70& 76.90& 49.30
\\
        Anaphor & self\_female & 86.60& 98.50& 86.70& 84.10& 42.00& 85.10
\\
        PolarityItem & any & 54.20& 85.60& 55.30& 52.70& 49.20& 54.60
\\
        PolarityItem & more\_or\_less & 20.20& 98.90& 46.80& 46.50& 46.70& 46.80
\\
        PolarityItem & even\_wh & 56.90& 92.40& 57.90& 53.60& 57.30& 55.90
\\
        DefinitenessEffect & definiteness\_every & 85.70& 94.60& 85.40& 83.30& 88.50& 84.20
\\
        DefinitenessEffect & definiteness\_demonstrative & 78.80& 96.20& 78.60& 75.20& 55.00& 77.30
\\
        Aspect & zai\_guo & 49.30& 97.30& 49.70& 47.90& 43.10& 49.20
\\
        Aspect & temporal\_le & 40.70& 63.40& 50.40& 49.10& 63.70& 50.30
\\
        Aspect & zai\_le & 49.80& 74.40& 49.90& 48.20& 69.00& 48.90
\\
        Aspect & temporal\_guo & 40.30& 88.10& 50.30& 47.60& 60.20& 50.10
\\
        Aspect & zai\_no\_le & 56.40& 77.90& 56.70& 53.80& 86.80& 55.20
\\
        WhFronting & mod\_wh & 54.70& 99.70& 54.40& 51.90& 36.10& 53.10
\\
        WhFronting & bare\_wh & 53.30& 100.00& 53.50& 50.30& 46.00& 52.40
\\
        Classifier-Noun & cl\_simple\_noun & 51.30& 98.00& 51.80& 49.70& 57.40& 50.70
\\
        Classifier-Noun & cl\_adj\_simple\_noun & 52.60& 96.30& 52.10& 50.10& 61.80& 51.30
\\
        Classifier-Noun & dem\_cl\_swap & 51.10& 99.60& 51.20& 49.20& 60.70& 50.60
\\
        Classifier-Noun & cl\_adj\_comp\_noun & 48.20& 70.60& 48.70& 46.90& 66.00& 47.50
\\
        Classifier-Noun & cl\_comp\_noun\_v2 & 49.60& 88.80& 49.30& 47.30& 61.90& 48.80
\\
        Classifier-Noun & cl\_comp\_noun & 51.00& 72.00& 51.60& 49.80& 61.30& 50.90
\\
        Classifier-Noun & cl\_adj\_comp\_noun\_v2 & 52.20& 89.50& 52.50& 50.70& 60.90& 52.10
\\
        AlternativeQuestion & haishi\_ma & 43.00& 95.00& 45.70& 45.90& 49.10& 45.70\\ \hline \hline 
        Average & & 51.32 & 83.41 & 56.23 & 54.27 & 58.79 & 55.48 \\ \hline \hline 
        \bottomrule
    \end{tabular}
\label{table:sling}
\end{table*}

\newpage

\begin{table*}[!ht]
\small
    \centering
    \caption{\textbf{(Japanese) } Accuracy of  the “vanilla” SSLM for Japanese (MAO-BabyBERTa) trained on CDS and the best performing objective curricula \textsc{+MMM} on each phenomenon in the Japanese Benchmark of Linguistic Minimal Pairs \cite{someya-oseki-2023-jblimp} compared to a Japanese monolingual GPT-2 LLM baseline trained on \(\approx 30B\) words and a \textsc{SSLM (wiki)} Baseline. }
    \begin{tabular}{lclcc}
        \hline \hline
        Phenomena & GPT2&  \textsc{Wiki}&Vanilla& MMM\\
        \hline
        Control/Raising & 16.67&  50.00&25.00 & 70.00 \\
        Island Effects & 75.76&  64.00&72.06& 92.19 \\
        Binding & 58.97&  79.05&57.86& 89.62 \\
        NPI Licensing & 50.00&  83.33&75.00 & 90.00 \\
        Argument Structure & 89.05&  41.6&54.82& 94.86 \\
        Ellipsis & 85.96&  49.36&56.13& 97.68 \\
        Verbal Agreement & 53.55&  57.82&69.22& 87.37 \\
        Filler-Gap & 55.56&  44.29&76.19& 85.71 \\
        Morphology & 82.86&  49.77&55.08& 82.05 \\
        Nominal Structure & 95.65&  41.51&55.87& 92.12 \\
        Quantifiers &  73.81&  48.96&60.56& 78.52 \\
        \hline
        Average & \textbf{77.95}&  \textbf{55.42}&\textbf{61.21} & \textbf{87.31} \\
        \hline \hline
    \end{tabular}

     \label{table:jblimp}
\end{table*}

\begin{table*}[!ht]
\small
    \centering
   \caption{(\textbf{French and German CLAMS) } Performance of \textsc{Growing, Inwards, MMM (upos)} in French and \textsc{MMM (upos)} in German (the best performing objective curricula) on CLAMS \cite{mueller-etal-2020-cross} compared to  \textsc{Mao-BabyBERTa} SSLM (“vanilla”) and the LLM and \textsc{SSLM (wiki)} baselines. We report the LLM baselines obtained by \citet{mueller-etal-2020-cross} for mBERT in French and German, which does not report results for “within objective relative” (object rel within) as all focus verbs for that particular language and construction were out-of-vocabulary. Chance CLAMS accuracy is 0.5. }
\begin{tabular}{|l|l|l|l|l|l|l|l|l|l|}
\hline
Language  &Model& Average & S-V& Obj Rel & Obj Rel & VP& Prep & Subject& Long VP\\ \hline 
 & & & & (within)& (across)& Coord& Animate& Relative&Coord\\\hline \hline
\textsc{French}  &LLM& 83.00\%& 100.00& – & 86.00& 100.00& 57.00& 57.00& 98.00\\ \hline
&\textsc{WIKI}& \(70.68\%\) & 67.48 & 73.40 & 73.80 & 71.27 & 66.80 & 70.80 & 71.27 \\ \hline
 &Vanilla& \(80.00\%\)& 82.0 & 64.90& 84.8 & 78.6 & 84.8 & 83.1 & 82.1 \\ \hline
 &Growing& \(76.21\%\) & 73.70 & 69.57 & 79.51 & 71.12 & 86.53 & 80.01 & 73.70 \\ \hline
 &Inwards& \(79.01\%\) & 76.95 & 68.50 & 84.10 & 75.86 & 83.80 & 87.00 & 76.89 \\ \hline
 &MMM& \(75.93\%\) & 82.33 & 72.60 & 74.40 & 81.79 & 65.80 & 70.90 & 83.71 \\ \hline
\textsc{German}  &LLM& 92.16\%& 95.00& – & 93.00& 97.00& 95.00& 73.00& 100.00\\ \hline
&\textsc{WIKI}& \(59.63\%\) & 56.55 & 47.90 & 60.60 & 55.32 & 57.20 & 60.60 & 79.28 \\ \hline
 &MMM& \(73.25\%\)& 75.32 & 79.80 & 66.40 & 78.52 & 68.40 & 66.40 & 77.90 \\ \hline
\end{tabular}
 \label{table:clams}
\end{table*}

\end{document}